\documentclass{article}
\usepackage[accepted]{synsml2023}

\usepackage{amsmath,amsfonts,bm}

\def\eqref#1{equation~\ref{#1}}

\def\1{\bm{1}}

\def\vb{{\bm{b}}}

\def\vn{{\bm{n}}}

\def\vx{{\bm{x}}}

\def\mW{{\bm{W}}}

\DeclareMathAlphabet{\mathsfit}{\encodingdefault}{\sfdefault}{m}{sl}
\SetMathAlphabet{\mathsfit}{bold}{\encodingdefault}{\sfdefault}{bx}{n}

\def\gX{{\mathcal{X}}}

\usepackage{xspace}

\usepackage{bm}

\usepackage[utf8]{inputenc} 
\usepackage[T1]{fontenc}    
\usepackage{url}            
\usepackage{booktabs}       
\usepackage{amsfonts}       
\usepackage{nicefrac}       
\usepackage{microtype}      
\usepackage{amsmath}
\usepackage{tikz}
\usepackage{multirow}
\usepackage{multicol}
\usepackage[colorlinks=true, allcolors=blue]{hyperref}     
\usepackage{cleveref}
\usepackage{amsthm}
\usepackage[symbol]{footmisc}
\usepackage[inline,shortlabels]{enumitem}

\crefname{algocf}{alg.}{algs.}
\Crefname{algocf}{Algorithm}{Algorithms}

\title{}
\begin{document}

\twocolumn[

\synsmltitle{INFINITY: Neural Field Modeling for Reynolds-Averaged Navier-Stokes Equations}
\synsmlsetsymbol{equal}{*}

\begin{synsmlauthorlist}
\synsmlauthor{Louis Serrano}{yyy}
\synsmlauthor{Leon Migus}{yyy,comp}
\synsmlauthor{Yuan Yin}{yyy}
\synsmlauthor{Jocelyn Ahmed Mazari}{sch}
\synsmlauthor{Patrick Gallinari}{yyy,zzz}
\end{synsmlauthorlist}

\synsmlaffiliation{yyy}{Sorbonne Université, CNRS, ISIR, 75005 Paris, France}
\synsmlaffiliation{comp}{Sorbonne Université, CNRS, Laboratoire Jacques-Louis Lions, 75005 Paris, France}
\synsmlaffiliation{sch}{Extrality, 75002 Paris, France}
\synsmlaffiliation{zzz}{Criteo AI Lab, Paris, France}

\synsmlcorrespondingauthor{Louis Serrano}{louis.serrano@isir.upmc.fr}
\synsmlcorrespondingauthor{Leon Migus}{leon.migus@isir.upmc.fr}

\synsmlkeywords{Machine Learning}

\vskip 0.3in
]

\printAffiliationsAndNotice%

\begin{abstract}
For numerical design, the development of efficient and accurate surrogate models is paramount. They allow us to approximate complex physical phenomena, thereby reducing the computational burden of direct numerical simulations. We propose INFINITY, a deep learning model  that utilizes implicit neural representations (INRs) to address this challenge. Our framework encodes geometric information and physical fields into compact representations and learns a mapping between them to infer the physical fields. We use an airfoil design optimization problem as an  example task and we evaluate our approach on the challenging AirfRANS dataset, which closely resembles real-world industrial use-cases. The experimental results demonstrate that our framework achieves state-of-the-art performance by accurately inferring physical fields throughout the volume and surface. Additionally we demonstrate its applicability in contexts such as design exploration and shape optimization: our model can correctly predict drag and lift coefficients while adhering to the equations.
\end{abstract}

\section{Introduction and motivation}

Numerical simulations are essential for analyzing systems governed by partial differential equations (PDEs) in fields like fluid dynamics and climate science. These simulations involve discretizing the domain and solving the equations using methods such as finite differences, finite elements, or finite volumes \citep{FEM, FDM, FVM}. Since direct numerical simulation (DNS) can be computationally expensive or intractable, it is crucial to develop computationally efficient yet accurate surrogate models to accelerate the design process.
Surrogate modeling for industrial applications, however, poses several challenges. The meshes used in these applications are extensive, consisting of hundreds of thousands of cells, and they also exhibit unstructured data and involve multi-scale phenomena.
A typical example is the design of airfoils which will be our application focus, although the ideas can be easily implemented for other design tasks. In this domain, a new costly simulation must be run for each mesh during the optimization process, leading to time-consuming processes. Additionally, the design process focuses on finding the optimal shape for an airfoil that minimizes the force required for flight. Experts typically maximize the lift-over-drag ratio by solving equations across the entire mesh, with particular emphasis on the surface where various multi-scale phenomena occur. 

Recently, deep learning methods have emerged as promising approaches for constructing surrogate models. However, the progress in this field was initially hindered by the lack of evaluation datasets representative of real-world data. The machine learning community has begun to address this issue by developing benchmarks.
In this work, we utilize the a recent AirfRANS dataset \cite{bonnet2022}, which aims to replicate real-world industrial scenarios. This comprehensive benchmark provides an evaluation framework to assess the capabilities of deep learning (DL) in modeling the two-dimensional incompressible steady-state Reynolds-Averaged Navier-Stokes (RANS) equations for airfoils. Additionally, this 2D dataset encompasses a wide range of airfoil shapes derived from NASA's early works
\citep{cummings2015applied}, various turbulence effects characterized by Reynolds numbers and different angles of attack.

The Navier-Stokes equations are widely used in fluid dynamics, and as a result, numerous neural network surrogates have been proposed for their modeling in different contexts. Initial attempts all relied on grid-based approaches such as convolutional Neural Networks (CNNs) \citep{um2020solver, thuerey2020deep, mohan2020embedding, wandel2020learning, obiols2020cfdnet, gupta2021multiwavelet, wang2020towards}. CNNs face challenges when dealing with the irregular meshes used in computational fluid dynamics (CFD). Graph Neural Networks (GNNs) have shown promise \cite{pfaff2020learning} but they have limitations in terms of receptive field size and information propagation across distant nodes, especially for large meshes. Additionally, GNNs struggle when the mesh is too dense and cannot fit into the memory of GPUs, necessitating sub-sampling. This limitation restricts their application in contexts where large meshes with multi-scale phenomena are prevalent. Furthermore, the evaluation of the models has primarily focused on traditional machine learning scores, such as global error over the entire domain (a.k.a. \textit{volume}), rather than more design-oriented scores, including local error in the surface area surrounding the airfoil (a.k.a. \textit{surface}) and errors in the aerodynamic forces of interest, such as drag and lift.

Leveraging recent advances in implicit neural representations (INRs) \citep{sitzmann2020implicit, mildenhall2021nerf}, which have shown successful applications in physics problems \citep{DiNO}, we introduce INFINITY, a model that utilizes coordinate-based networks to encode geometric information and physical fields into concise representations. INFINITY establishes a mapping between variables representing the problem's geometry and the corresponding physical fields, within this representation space. It possesses several unique features: (i) it is robust to varying mesh sampling, allowing for adaptability to different geometries, (ii) it effectively captures multi-scale phenomena, resulting in state-of-the-art scores for both volume and surface evaluations, (iii) as a continuous surrogate model, it can be used to accelerate the evaluation of different meshes during the design process, leading to significant speed-up. 
Importantly, we verified that INFINITY's field predictions accurately produce the correct lift and drag forces clearly outperforming all the baselines.

\section{Method}

\begin{figure*}
    \centering
    \includegraphics[width=1.0\linewidth]{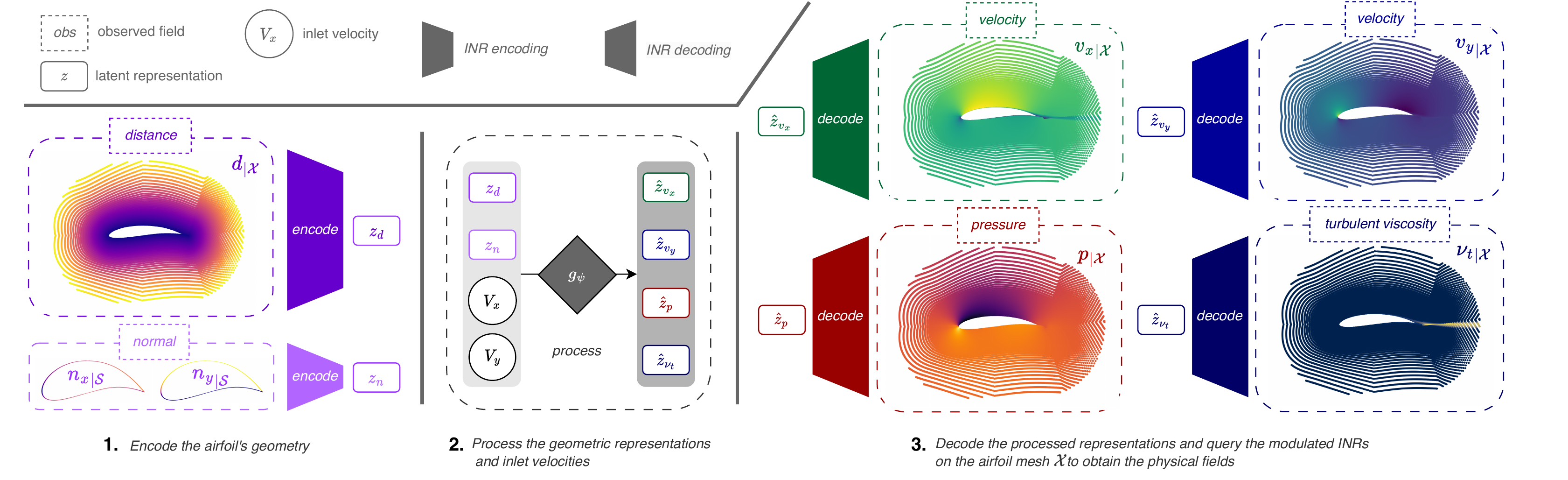}
    \caption{The inference of INFINITY proceeds in three steps. 1. We encode the distance function $d$ and the normal components $n_x, n_y$ into the latent representations $z_d$ and $z_n$. 2. We process these codes along with the inlet velocities $V_x,  V_y$ to obtain the predicted output codes $\hat{z}_{{v_x}}, \hat{z}_{{v_y}}, \hat{z}_{{p}}, \hat{z}_{{\nu_t}}$ corresponding respectively to velocity, pressure and viscosity. 3. The processed codes are decoded with the modulated INRs, which can be queried directly at any mesh position $\vx \in \gX$.}
    \label{fig:inference}
\end{figure*}

\subsection{Problem setting} \label{sec:pb_setting}
We aim at proposing a surrogate model for airfoil design optimization in scenarios where the amount of available training data is limited ($n_{tr} \leq 1000$). Each airfoil is associated with a domain $\Omega_i$, which is linked to a specific geometry. Consequently, different meshes $\mathcal{X}_i$ are generated within each domain. The characterization of an airfoil involves defining boundary conditions on $\partial\Omega_i$ corresponding to the airfoil surface, which are discretized into a surface mesh $\mathcal{S}_i$.

The geometric inputs for our model include the following information: \begin{itemize*}
    \item \textit{Node positions $\vx$} represent the coordinates of each node within the airfoil's domain.
    \item \textit{Distance function $d(\vx)$} provides the distance from each node to the surface of the airfoil.
    \item \textit{Normal vectors of the mesh nodes on the airfoil surface $\vn(\vx) = (n_x(\vx), n_y(\vx))$} specify the direction perpendicular to the airfoil surface at each node.
\end{itemize*}
In addition to the geometric inputs, we also have access to the inlet velocity values $V_x$ and $V_y$, denoting the horizontal and vertical components of the velocity, respectively. It is worth noting that, on average, a mesh consists of approximately 200,000 nodes, providing a detailed representation of the airfoil's geometry.

The primary objective of the design optimization process is to maximize the lift-over-drag coefficient ratio, which serves as the key performance metric. To achieve this, we place significant emphasis on evaluating the relative errors in both the drag and lift coefficients, as well as assessing the Spearman correlation between predicted and actual values.

Rather than directly predicting the drag and lift values, our approach focuses on inferring various fluid fields associated with the airfoil's geometry. This includes calculating the velocities $(v_x, v_y)$, pressure $p$, and turbulent kinematic viscosity $\nu_t$ on the mesh nodes, following the experimental protocol proposed in \cite{bonnet2022}. Therefore the inputs of our surrogate model are $(V_x, V_y, d|_{\mathcal{X}_i}, {n_x}|_{\mathcal{S}_i}, {n_y}|_{{\mathcal{S}_i}})_{i=1}^{n_{tr}}$, and the outputs are 
$({v_x}|_{\mathcal{X}_i}, {v_y}|_{\mathcal{X}_i}, {p}|_{\mathcal{X}_i}, {\nu_t}|_{\mathcal{X}_i})_{i=1}^{n_{tr}}$.
The output physical fields provide valuable insights into the underlying behavior of the fluid and its interaction with the airfoil's geometry. The drag and lift coefficients are calculated based on the predictions of the trained model while respecting the form of the RANS equations. This approach enables us to obtain a comprehensive understanding of the underlying fluid behavior and its relationship with the airfoil's geometry, thereby ultimately enhancing the accuracy of drag and lift estimation.

\subsection{Model}

We present INFINITY: Implicit Neural Fields for INterpretIng geomeTry and inferring phYsics. 

\paragraph{Modulated INR}

In our model, we will treat each geometric input ($d$ or $n$) or physical output function ($v, p$, or $\nu$) separately and each will be modeled by an INR. Let us then consider a generic function $u$, which will represent either an input geometric field or an output physical field defined over a domain $\Omega$ or at its boundary $\partial \Omega$. Let us denote $u_i$ the function corresponding to a specific airfoil example. $u_i$ will be represented by an INR $f_{\theta_u,\phi_{u_i}}$ with two sets of parameters: parameters $\theta_u$ shared by all the $u_i$, and modulation parameters $\phi_{u_i}$ specific to each individual function $u_i$. In our airfoil example, $\phi_{u_i}$ enables the INR to handle different geometries. Overall, this decomposition allows the modulated INR to capture both shared characteristics among the example's functions $u_i$ and the unique properties of each one. INFINITY leverages latent representations inferred from the modulation spaces of the INRs. These latent representations, denoted as $z_{u_i}$, are compact codes that encode information from the INRs' parameters. They serve as inputs to a hypernetwork $h_u$, with weights $w_u$, which computes the modulation parameters $\phi_{u_i} = h_u(z_{u_i})$. 

In this work we use Fourier Features \citep{Tancik2020} as an INR backbone and apply shift modulation \citep{Perez2018}: $f_{\theta, \phi_{u_i}}(\vx) = \mW_L\bigl(\chi_{L-1} \circ \chi_{L-2} \circ \cdots  \circ \chi_0 (\vx)\bigr) + \vb_L$, with $\chi_j(\eta_j) = \sigma\bigl(\mW_j \eta_j + \vb_j +  (\bm{\phi_{u_i}})_j\bigr)$.  We note $\eta_0 = \vx$ and $(\eta_j)_{j\geq 1}$ the hidden activations throughout the network. Hence, the parameters $\theta = (\mW_j, \vb_j)_{j=0}^{L}$ are shared between all examples and the modulation $\phi_{u_i} = ((\bm{\phi_{u_i}})_j)_{j=0}^{L-1}$ is specific to a single example.
We compute the modulation parameters $\phi_{u_i} = ((\bm{\phi_{u_i}})_j)_{j=0}^{L-1}$ from $z$ with a linear hypernetwork.

With the learned shared parameters $(\theta_u, w_u)$, the modulated INR enables two processes: decoding and encoding (see \Cref{fig:inference}). Decoding refers to mapping a given code $z_{u_i}$ to the corresponding INR function $f_{\theta_u, \phi_{u_i}}$, where $\phi_{u_i} = h_u(z_{u_i})$, while encoding involves generating a code $z_{u_i}$ given a function $u_i$, providing a compact representation of the function within the modulation space of the INR. 

To obtain the compact code $z_{u_i}$ for reconstructing the original field $u_i$ using the INR, an inverse problem is solved through a procedure called \textit{auto-decoding}. The objective is to compress the necessary information into $z_{u_i}$ such that the reconstructed value $\tilde{u}_i(\vx)=f_{\theta_u,\phi_{u_i}}(\vx)$ approximates the original value $u_i(\vx)$ for all $\vx \in \mathcal{X}_i$. 
The approximate solution to this inverse problem is computed iteratively through a gradient descent optimization process:
\begin{equation}
\label{eqn-encoding}
\begin{aligned}
&z_{u_i}^{(0)} = 0, \\
&z_{u_i}^{(k+1)} = z_{u_i}^{(k)} -\alpha \nabla_{z_{u_i}^{(k)}} \mathcal{L}_{\mu_i}(f_{\theta_u, \phi^{(k)}_{u_{i}}}, u_i), \\
&\text{with} \ \phi^{(k)}_{u_i} = h_u(z_{u_i}^{(k)}) \ \text{for } 0 \leq k \leq K - 1.\\
\end{aligned}
\end{equation}
where $\alpha$ is the inner loop learning rate, $K$ the number of inner steps, and $\mathcal{L}_{\mu_i}(u_i, \tilde{u}_i) = \mathbb{E}_{\vx \sim \mu_i}[(u_i(\vx) - \tilde{u}_i(\vx))^2]$ where $\mu_i$ is a measure defined through the observation grid $\mathcal{X}_i$  $\mu_i(\cdot) = \sum_{x\in \mathcal{X}_i} \delta_x(\cdot)$, with $\delta_x(\cdot)$ the Dirac measure.

As indicated before, we treat each input and output function independently: there are two input functions denoted as $(d, n)$ and four output functions denoted as $(v_x, v_y, p, \nu_t)$. Each $u_i \in \{d, n, v_x, v_y, p, \nu_t\}$ is represented by a modulated INR $f_{\theta_u,\phi_{u_i}}$, where $u_i$ stands for a field specific to an airfoil example. INFINITY then learns a mapping between the latent representations of the geometric input fields and the latent representations of the physics output fields.

\paragraph{Inference} As illustrated in Figure \ref{fig:inference}, INFINITY follows a three-step procedure: \textit{encode}, \textit{process}, and \textit{decode}.
\begin{itemize}
\item \textit{Encode:} Given the geometric input functions $d_i, n_i$ and the corresponding INR learned parameters, respectively $\theta_{d}, w_d$ and $\theta_{n}, w_n$, functions $d_i, n_i$ are encoded into the latent codes $z_{d_i}, z_{n_i}$ according to \Cref{eqn-encoding}. Since we can query the INRs anywhere within the domain, we can hence freely encode functions without mesh constraints. This lets us freely encode inputs with different geometries.
\item \textit{Process:} Once we obtain $z_{d_i}$ and $z_{n_i}$, we can infer the latent output codes $\big(\hat{z}_{{v_x}_i}, \hat{z}_{{v_y}_i}, \hat{z}_{{p}_i}, \hat{z}_{{\nu_t}_i}\big) = g_\psi\bigg(\big(z_{d_i}, z_{n_i}, {V_x}_i, {V_y}_i\big)\bigg)$. We consider here that $g_\psi$ is implemented through an MLP with parameters $\psi$. 
\item \textit{Decode:} We decode each processed output code $\big(\hat{z}_{{v_x}_i}, \hat{z}_{{v_y}_i}, \hat{z}_{{p}_i}, \hat{z}_{{\nu_t}_i}\big)$ with their associated hypernetwork and modulated INR. We make use of the INRs to freely query a physical field at any point within its spatial domain. These components generate the final output functions by mapping the latent codes back to the output space.
\end{itemize}

\begin{table*}[ht]
    \centering
    \resizebox{\linewidth}{!}{
    \begin{tabular}{ccccccc}     
    \toprule
    	& & INFINITY & GraphSAGE & MLP  & Graph U-Net & PointNet \\
    \midrule
    
    \multirow{4}{*}{Volume} & $v_x$	& \textbf{0.06 $\pm$ 0.01} & 0.83 $\pm$ 0.01  & 0.95 $\pm$ 0.06  & 1.52 $\pm$ 0.34 & 3.50 $\pm$ 1.04 \\
    & $v_y$	&   \textbf{0.06 $\pm$ 0.01} & 0.99 $\pm$ 0.05 & 0.98 $\pm$ 0.17 & 2.03 $\pm$ 0.39 & 3.64 $\pm$ 1.26 \\
    & $p$	&  \textbf{0.25 $\pm$ 0.01} & 0.66 $\pm$ 0.05 & 0.74 $\pm$ 0.13 & 0.66 $\pm$ 0.08 & 1.15 $\pm$ 0.23  \\
    & $\nu_t$	&   \textbf{1.32 $\pm$ 0.08}  & 1.60 $\pm$ 0.21 & 1.90 $\pm$ 0.10  & 1.46 $\pm$ 0.14 & 2.92 $\pm$ 0.48 \\
    \midrule      
    Surface & $p_{|\mathcal{S}}$	&  \textbf{0.07 $\pm$ 0.01}  & 0.66 $\pm$ 0.10  & 1.13 $\pm$ 0.14 & 0.39 $\pm$ 0.07 & 0.93 $\pm$ 0.26 \\
    \midrule
    
    \multirow{2}{*}{Relative error} & $C_D$ &  \textbf{0.366 $\pm$ 0.023} & 4.050 $\pm$ 0.704  & 4.289 $\pm$ 0.679 & 10.385 $\pm$ 1.895 & 14.637 $\pm$ 3.668 \\
    & $C_L$ &  \textbf{0.081 $\pm$ 0.007} & 0.517 $\pm$ 0.162  & 0.767 $\pm$ 0.108 & 0.489 $\pm$ 0.105 & 0.742 $\pm$ 0.186 \\
    \cmidrule(lr){2-7}
    \multirow{2}{*}{Spearman correlation} & $\rho_D$ &  \textbf{0.578 $\pm$ 0.050} & -0.303 $\pm$ 0.124  & -0.117 $\pm$ 0.256 & -0.138 $\pm$ 0.258 & -0.022 $\pm$ 0.097 \\
    & $\rho_L$ & \textbf{0.997 $\pm$ 0.001}  & 0.965 $\pm$ 0.011  & 0.913 $\pm$ 0.018 & 0.967 $\pm$ 0.019 & 0.938 $\pm$ 0.023 \\
    \midrule

    Inference time ($\mu$s) & & 98 $\pm$ 70  & 20.9 $\pm$ 2.3 & 13.3 $\pm$ 0.2 & 357.8 $\pm$ 36.9  & 33.9 $\pm$ 3.5 \\

    \bottomrule

    \end{tabular}}
    \caption{Test results on AirfRANS. Mean squared error (MSE) on normalized fields expressed with factor $(\times 10^{-2})$ for the volume and $(\times 10^{-1})$ for the surface. Relative errors  $C_D, C_L$ on the drag and lift and Spearman correlations $\rho_D, \rho_L$ on the drag and lift. The results from the baselines are taken from \citet{bonnet2022}.} 
    \label{tab:inr_baselines_full}
\end{table*}

\subsection{Training}
\label{section-training}

We implement a two-step training procedure that first learns the modulated INR parameters $\theta_u$ and $\phi_{u_i}$ for all input and output functions, before training the map $g_\psi$. During the training of the INRs we force the \textit{auto-decoding} process to take only a few gradient steps to encode the geometric functions or physical fields. This enhances the INR capability to encode new geometrical inputs in a few steps at test time, and also reduces the space size of the target output codes. This regularization prevents the different INRs to memorize the training sets into the individual codes. In order to obtain a network that is capable of quickly encoding new geometrical and physical inputs, we employ a second-order meta-learning training algorithm based on CAVIA \citep{Zintgraf2018}. Compared to a first-order scheme such as Reptile \citep{Reptile}, the outer loop back-propagates the gradient through the $K$ inner steps, consuming more memory. Indeed, we need to compute gradients of gradients but this yields higher reconstruction results with the modulated INR. We experimentally found that using 3 inner-steps for training, or testing, was sufficient to obtain very low reconstruction errors for the geometric or physical fields. Using more inner-steps would result in a higher computation cost with only a marginal gain in reconstruction capacity. We outline the training pipeline of a modulated INR in \Cref{alg:inr-input-training}.
 \begin{algorithm}
 \caption{Modulated INR training}
 \label{alg:inr-input-training}
 \begin{algorithmic}[1]

     \WHILE{convergence is false}
         \STATE \textit{Sample a batch} $\mathcal{B}$ \textit{of data} $(u_i)_{i \in \mathcal{B}}$
         \STATE \textit{Set codes to zero:} $z_{u_i} \gets 0$ for $i$ in $\mathcal{B}$
         \STATE \textit{Perform input encoding inner loop:}
         \FOR{$i$ in $\mathcal{B}$ and step in $\{1, ..., K_u\}$}
             \STATE \quad $\phi_{u_i} = h_u(z_{u_i})$
             \STATE \quad $z_{u_i} \gets z_{u_i} - \alpha_a \nabla_{z_{u_i}} \mathcal{L}_{\mathcal{X}_{i}}(f_{\theta_u, \phi_{u_i}}, u_i)$
         \ENDFOR
         \FOR{$i$ in $\mathcal{B}$:}
             \STATE \quad $\phi_{u_i} = h_u(z_{u_i})$
         \ENDFOR
         \STATE \textit{Perform outer loop update:}
         \STATE $\theta_u \gets \theta_u - \eta \frac{1}{|\mathcal{B}|} \sum_{i \in \mathcal{B}} \nabla_{\theta_u} \mathcal{L}_{\mathcal{X}_{i}}(f_{\theta_u, \phi_{u_i}}, u_i)$
         \STATE $w_u \gets w_u - \eta \frac{1}{|\mathcal{B}|} \sum_{i \in \mathcal{B}} \nabla_{w_u} \mathcal{L}_{\mathcal{X}_{i}}(f_{\theta_u, \phi_{u_i}}, u_i)$
     \ENDWHILE
 \end{algorithmic}
 \end{algorithm}
  Once the different INRs have been fitted, we encode the functions into the input codes $z_{d_i}, z_{n_i}$ and target codes $z_{{v_x}_i}, z_{{v_y}_i}, z_{{p}_i}, z_{{\nu_t}_i}$. The training of $g_\psi$ is performed in the small dimensional $z$-code space, and is supervised through the MSE loss with the target codes.

\section{Experiments}
\paragraph{Baselines}
We use the same baselines as \citet{bonnet2022};  GraphSAGE \citep{GraphSAGE}, a PointNet \citep{pointnet}, a Graph U-Net \citep{graphunet} and a MLP. Those baselines have been initially chosen as they process in different ways the inputs. The results are given for the setup ``full data regime'' of AirfRANS, using 800 samples for training and 200 for testing.
\paragraph{Results}

In \Cref{tab:inr_baselines_full}, the INFINITY model demonstrates superior inference capabilities on the volume and surface compared to the baselines. Indeed, It achieves significantly lower error values on the volume velocity and pressure fields, while exhibiting an order-of-magnitude lower MSE on the surface pressure. This substantial gain in prediction power translates to order of magnitude lower relative errors on the drag and lift forces, accompanied by high positive Spearman correlations. These results indicate a strong alignment between INFINITY's predictions and the true drag and lift forces. Consequently, INFINITY emerges as the only model capable of predicting accurately physical fields on the volume and surface while maintaining coherent and accurate drag and lift estimations.
On the downside, the INFINITY model has a longer inference time compared to GraphSAGE and PointNet. However, this increased inference time is still within an acceptable range, considering its superior performance and that a numerical solver needs approximately $20$ minutes to complete a simulation. Furthermore, it is counterbalanced by the ability to query the full mesh directly, in stark contrast to graph-based methods that necessitate sub-sampling to process the inputs.

\section{Conclusion}

We introduce INFINITY, a model that utilizes coordinate-based networks to encode geometric information and physical fields into compact representations. INFINITY establishes a mapping between geometry and physical fields within a reduced representation space. We validated our model on AirfRANS, a challenging dataset for the Reynolds-Averaged Navier-Stokes equation, where it significantly outperforms previous baselines across all relevant performance metrics. At post-processing stage, the predicted fields yield accurate lift and drag forces. This validates INFINITY's potential as a surrogate design model, where it could be plugged in any design optimization or exploration loop.

\section*{Broader impact}

This work could serve multiple purposes, including:

\begin{itemize}
    \item Enhancing surrogate solvers for Computational Fluid Dynamics (CFD) engineers, facilitating efficient design iterations.
    \item Cost and risk reduction in prototyping new plane designs, mitigating real-world study expenses.
\end{itemize}

\section*{Acknowledgements} We acknowledge the financial support provided by DL4CLIM (ANR-19-CHIA-0018-01) and DEEPNUM (ANR-21-CE23-0017-02) ANR projects, as well as the program France 2030, project 22-PETA-0002.

\bibliographystyle{abbrvnat}
\bibliography{bibliography}

\end{document}